\pdfoutput=1

\documentclass[11pt]{article}

\usepackage{tabularx, makecell} 
\usepackage[]{acl}
\usepackage{times}
 \usepackage{booktabs}
\usepackage{amsmath}
\usepackage{bm}
\usepackage[T1]{fontenc}
\usepackage[utf8]{inputenc}
\usepackage{microtype}
\usepackage{inconsolata}
\usepackage{graphicx}
\usepackage{comment}
\usepackage{url}
\usepackage{cleveref}
\usepackage{xspace}
\usepackage{hyphenat}
\usepackage{makecell}

\usepackage[utf8]{inputenc}
\usepackage[T1]{fontenc}

\usepackage{amssymb}              
\newcommand{\cmark}
{\ensuremath{\checkmark}}  
\newcommand{\xmark}{\ensuremath{\times}}
\newcommand{\snum}[2]{\ifx#1+\mathord{+}\else\mathord{-}\fi #2}
\usepackage{soul}

\usepackage{array}     
\usepackage{adjustbox} 
\usepackage{booktabs}
\usepackage{subcaption}  

\title{\textsc{Merlin}: Multi-Stage Curriculum Alignment for Multilingual Encoder-LLM Integration in Cross-Lingual Reasoning}

\author{
  Kosei Uemura$^{1}$ \quad
  David Guzmán$^{2}$ \quad
  Quang Phuoc Nguyen$^{3}$ \quad
  Jesujoba Oluwadara Alabi$^{4}$ \\
  \textbf{En-Shiun Annie Lee}$^{3,1}$ \quad
  \textbf{David Ifeoluwa Adelani}$^{2,5}$ \\
  \\
  $^{1}$University of Toronto \quad
  $^{2}$Mila-Quebec AI Institute, McGill University \quad \\
  $^{3}$OntarioTech University
  $^{4}$Saarland University \quad
 $^{5}$Canada CIFAR AI Chair \\
  \texttt{\{k.uemura,enshiun.lee\}@mail.utoronto.ca} \\
  \texttt{quangphuoc.nguyen@ontariotechu.net} \\
  \texttt{jalabi@lsv.uni-saarland.de} \\
  \texttt{\{david.guzman, david.adelani\}@mila.quebec}
}

\begin{document}
\maketitle
\begin{abstract}

Large language models (LLMs) excel in English but still struggle with complex reasoning in many low-resource languages (LRLs). Existing methods align LLMs with multilingual encoders, such as LangBridge and MindMerger, raising the accuracy for mid and high-resource languages, yet large performance gap remains for LRLs. We present \textsc{Merlin}, a model-stacking framework that iteratively refines in two-stages based on a curriculum strategy (from general to specific where general is bilingual bitext and specific is task-specific data) and adapts only a small set of DoRA weights. On the AfriMGSM benchmark \textsc{Merlin} improves exact-match accuracy by +12.9 pp over MindMerger and outperforms GPT-4o-mini by 15.2 pp. It also yields consistent gains on MGSM and MSVAMP (+0.9 and +2.8 pp), demonstrating effectiveness across both low and high-resource settings. \footnote{The implementation of \textsc{Merlin} is publicly available at \url{https://github.com/LLMforLRL/MERLIN}.}

\end{abstract}

\section{Introduction}

Large language models (LLMs) have achieved impressive results across a broad range of natural language processing (NLP) tasks \citep{openai2024gpt4technicalreport, jiang2024mixtralexperts, geminiteam2024gemini15unlockingmultimodal}. Despite these advances, their performance often deteriorates significantly when dealing with \emph{low-resource languages} (LRLs) \citep{adelani2024irokobenchnewbenchmarkafrican, uemura-etal-2024-afriinstruct, ojo2024goodlargelanguagemodels}. This shortfall is partly due to the limited online presence of LRLs both in data availability and NLP tools, and heavy focus on high or mid-resource languages during the pretraining of these models \citep{touvron2023llama2openfoundation,ranathunga_neural_2023,aryabumi2024aya}.

To improve reasoning capabilities for LRLs, the most straightforward approach is to adapt LLMs through continual pre-training~\citep{alves2024tower,ng2025sea,buzaaba2025lugha} or post-training~\citep{ogundepo2025improving}. However, these methods often depend on access to in-domain data, which is scarce for LRLs due to the lack of annotators, speakers, and linguistic experts. Additionally, fine-tuning LLMs---especially those with billions of parameters---incurs substantial computational costs and energy usage, making these approaches less feasible in resource-constrained settings \citep{dettmers2023qloraefficientfinetuningquantized,han2024parameterefficientfinetuninglargemodels}. LangBridge \citep {yoon-etal-2024-langbridge} and MindMerger \citep{huang2024mindmergerefficientboostingllm} address this challenge by leveraging rich multilingual representations from an external encoder as input to decoder-only LLMs. This design significantly enhances multilingual reasoning capabilities while substantially reducing both data and computational requirements. 
On the other hand, curriculum learning---a training paradigm in which models are exposed to data of increasing difficulty---has been shown to improve both convergence speed and final performance \protect\citep{soviany_curriculum_2022} and has been applied in recent works to enhance LLMs reasoning ability \protect\citep{ma-etal-2025-problem}.

\begin{figure*}[htp]
    \centering
    \includegraphics[width=\textwidth]{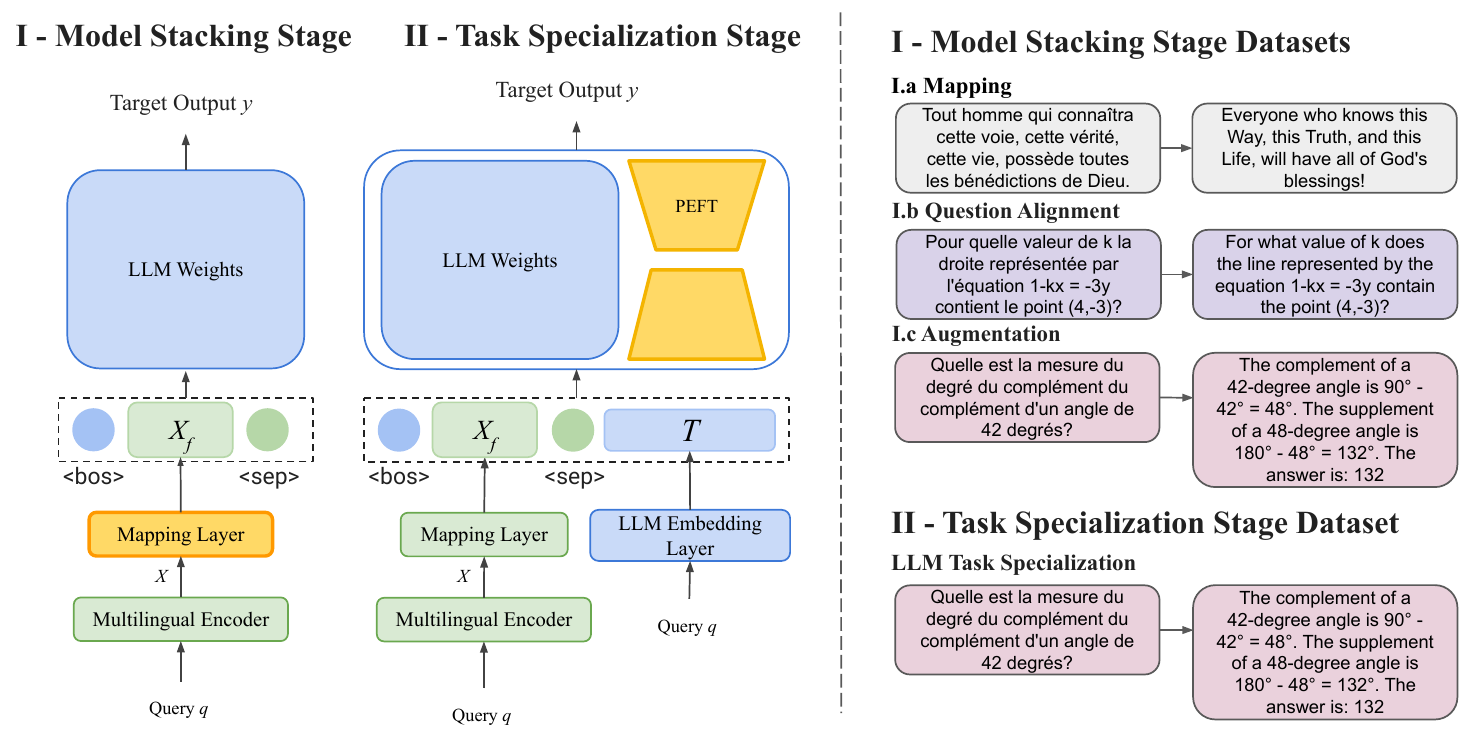}
    \caption{Overview of \textsc{Merlin} (\textbf{M}ultilingual \textbf{E}mbedding-Enhanced \textbf{R}easoning for \textbf{L}anguage \textbf{I}ntegration \textbf{N}etwork), our two-stage framework for multilingual reasoning. In the Mapping Stage, a lightweight mapping layer is trained sequentially on three datasets—starting with general bilingual translation, followed by query-pair translation, and finally task-specific QA translation—to align the multilingual encoder’s outputs with the LLM’s embedding space. In the Embedding Enhancement Stage, this mapping “connector” remains frozen while the LLM body is fine-tuned via parameter-efficient PEFT on QA data, thereby strengthening cross-lingual reasoning.}
    \label{fig:mindmerger-ee}
    \vspace*{-1em}
\end{figure*}

In this paper, we introduce \textbf{M}ultilingual \textbf{E}mbedding-Enhanced \textbf{R}easoning for \textbf{L}anguage \textbf{I}ntegration \textbf{N}etwork (Figure~\ref{fig:mindmerger-ee}), which is abbreviated to \textbf{\textsc{Merlin}}.  It is a two-stage framework that integrates insights from model stacking with question alignment. This design combines data-efficient supervised training with targeted adaptation of the LLM’s internal representations, yielding tighter encoder–decoder alignment without the cost of full-model retraining.  Inspired by the principles of curriculum learning, in the first stage, we learn a lightweight mapping layer to align a strong multilingual encoder with a frozen LLM through three successive tasks starting with general domain bilingual data, followed by math domain bilingual data, and finally the task data. In the second stage, we fine-tune the LLM with parameter-efficient methods while keeping the mapping layer fixed, enabling the LLM to internalize and better exploit the learned representations during reasoning.
We evaluate \textsc{Merlin} on various math reasoning datasets such as Multilingual Grade School Math Benchmark (MGSM) \citep{shi2022languagemodelsmultilingualchainofthought}, Multilingual Semantic VAriant Math Problems (MSVAMP) \citep{chen2024breakinglanguagebarriersmultilingual}, and AfriMGSM~\citep{adelani2024irokobenchnewbenchmarkafrican}, an extension of MGSM to several low-resource African languages. \textsc{Merlin} surpasses the strongest baseline, MindMerger, LayAlign, SLAM and QAlign, by more than \textbf{+11.0} accuracy points on AfriMGSM and delivers consistent gains on MGSM with \textbf{+2.8} performance gain to MindMerger, demonstrating its effectiveness for cross-lingual mathematical reasoning in multilingual settings, including LRLs. 
To assess generalization beyond mathematical reasoning, we evaluate \textsc{Merlin} on natural language inference tasks using AfriXNLI~\citep{adelani2024irokobenchnewbenchmarkafrican}, with +1.6 improvement in performance over MindMerger. 

\begin{table*}[ht]
    \centering
    \tiny
    \setlength{\tabcolsep}{4pt}
    \renewcommand{\arraystretch}{0.8}
    \resizebox{\textwidth}{!}{
    \begin{tabular}{@{}lcccccccccc|ccc@{}}
        \toprule
        \textbf{Model} & \textbf{Bn} & \textbf{Th} & \textbf{Sw} & \textbf{Ja} &
        \textbf{Zh} & \textbf{De} & \textbf{Fr} & \textbf{Ru} &
        \textbf{Es} & \textbf{En} & \textbf{MRL} & \textbf{HRL} & \textbf{Avg} \\
        \midrule
        \multicolumn{14}{l}{\textbf{MGSM}}\\
        \midrule
        \multicolumn{14}{l}{\textit{Open Models}}\\
        Metamath 7B   &  6.4 &  2.0 &  4.4 & 36.4 & 40.4 & 52.8 & 50.4 & 46.0 & 58.4 & 63.6 &  4.3 & 49.7 & 36.1 \\
        Gemma 2 9B    & 56.4 & 66.4 & 64.4 & 65.2 & 69.6 & 70.4 & 59.2 & 67.6 & 72.0 & 76.0 & 62.4 & 68.6 & 66.7 \\
        \midrule
        \multicolumn{14}{l}{\textit{Key Multilingual Approaches–Metamath 7B \& mT5‑XL}}\\
        LangBridge    & 42.8 & \textbf{50.4} & 43.2 & 40.0 & 45.2 & 56.4 & 50.8 & 52.4 & 58.0 & 63.2 & 45.5 & 52.3 & 50.2 \\
        QAlign-LoRA        & 32.4 & 39.6 & 40.4 & 44.0 & 48.4 & 54.8 & 56.8 & 52.4 & 59.6 & \textbf{68.0} & 37.5 & 54.9 & 49.6 \\
        SLAM          & 32.0 & 44.0 & 40.0 & 46.0 & 48.4 & 54.0 & 55.2 & 54.8 & 56.8 & 64.8 & 38.7 & 54.3 & 49.6 \\
        LayAlign      & 1.6 & 3.6 & 8.0 & 23.2 & 23.2 & 37.2 & 37.6 & 34.8 & 38.4 & 43.2 & 4.4 & 33.9 & 25.1 \\
        MindMerger    & \textbf{43.6} & 44.4 & 42.5 & 45.6 & 45.6 & \textbf{56.8} & \textbf{58.0} & 56.0 & 54.0 & 64.0 & 43.5 & 54.3 & 51.1 \\
        \textsc{Merlin} & 43.2 & 49.2 & \textbf{44.8} & \textbf{48.4} & \textbf{48.8} & 53.2 & 55.6 & \textbf{58.0} & \textbf{59.2} & 63.6 & \textbf{45.7} & \textbf{55.3} & \textbf{52.4} \\
        $\Delta$ (\textsc{Merlin}–MM) &
           $\snum{-}{0.4}$ & $\snum{+}{4.8}$ & $\snum{+}{2.3}$ & $\snum{+}{2.8}$ & $\snum{+}{3.2}$ &
           $\snum{-}{3.6}$ & $\snum{-}{2.4}$ & $\snum{+}{2.0}$ & $\snum{+}{5.2}$ & $\snum{-}{0.4}$ &
           $\mathbf{\snum{+}{2.2}}$ & $\mathbf{\snum{+}{1.0}}$ & $\mathbf{\snum{+}{1.3}}$ \\
        \midrule
        \multicolumn{14}{l}{\textit{Key Multilingual Approaches–Gemma 2 9B \& NLLB‑600M‑distilled}}\\
        QAlign-LoRA & 58.0 & 60.8 & 62.0 & 56.4 & 62.4 & 68.8 & 68.0 & 68.0 & 76.8 & 80.8 & 60.3 & 68.7 & 66.2\\
        SLAM          & 63.2 & 61.2 & 52.4 & 62.4 & 64.4 & 68.8 & 53.6 & 70.0 & 69.2 & 69.2 & 58.9 & 65.4 & 63.4 \\
        MindMerger    & 66.8 & 72.8 & 69.2 & 66.8 & 72.8 & 78.0 & 75.2 & 73.6 & 76.4 & 82.4 & 69.6 & 75.0 & 73.4 \\
        \textbf{\textsc{Merlin}}& \textbf{74.8} & \textbf{76.0} & \textbf{77.6} & \textbf{71.6} & \textbf{72.8} &
                         73.2 & 74.8 & \textbf{78.4} & \textbf{77.6} & \textbf{85.6} &
                         \textbf{76.1} & \textbf{76.3} & \textbf{76.2} \\
        $\Delta$ (\textsc{Merlin}–MM) &
           $\snum{+}{8.0}$ & $\snum{+}{3.2}$ & $\snum{+}{8.4}$ & $\snum{+}{4.8}$ & 0.0 &
           $\snum{-}{4.8}$ & $\snum{-}{0.4}$ & $\snum{+}{4.8}$ & $\snum{+}{1.2}$ & $\snum{+}{3.2}$ &
           $\mathbf{\snum{+}{6.5}}$ & $\mathbf{\snum{+}{1.3}}$ & $\mathbf{\snum{+}{2.8}}$ \\
        \midrule
        \multicolumn{14}{l}{\textbf{MSVAMP}}\\
        \midrule
        \multicolumn{14}{l}{\textit{Open Models}}\\
        Metamath 7B   & 11.8 & 13.1 & 11.8 & 40.9 & 48.2 & 56.1 & 58.0 & 54.8 & 59.5 & 63.0 & 12.2 & 54.4 & 41.7 \\
        Gemma 2 9B    & 63.9 & 72.0 & 71.2 & 78.7 & 81.8 & 78.7 & 79.3 & 78.3 & 80.3 & 80.7 & 69.0 & 79.7 & 76.5 \\
        \midrule
        \multicolumn{14}{l}{\textit{Key Multilingual Approaches–Metamath 7B \& mT5‑XL}}\\
        LangBridge    & 46.8 & 46.3 & 42.1 & 45.5 & 50.4 & 58.1 & 57.0 & 55.8 & 56.9 & 60.6 & 45.1 & 54.9 & 52.0 \\
        QAlign-LoRA        & 41.7 & 47.7 & 54.8 & 58.0 & 55.7 & 62.8 & 63.2 & 61.1 & 63.3 & 65.3 & 48.1 & 61.3 & 57.4 \\
        SLAM          & 49.1 & 50.6 & 55.4 & 60.6 & 63.1 & 64.6 & 65.4 & \textbf{64.1} & \textbf{66.6} & 65.3 & 51.7 & 64.2 & 60.5 \\
        LayAlign      & 7.3 & 15.4 & 16.9 & 41.7 & 45.9 & 56.4 & 55.3 & 52.9 & 55.0 & 59.3 & 13.2 & 52.4 & 40.6 \\
        MindMerger    & 51.1 & 47.9 & 53.0 & 54.7 & 53.6 & 58.4 & 60.7 & 57.7 & 60.5 & 64.1 & 50.7 & 58.5 & 56.2 \\
        \textbf{\textsc{Merlin}}& \textbf{53.1} & \textbf{55.6} & \textbf{58.9} & \textbf{62.0} & \textbf{62.3} & \textbf{65.5} & \textbf{65.7} & 62.2 & 64.9 & \textbf{68.2} & \textbf{55.9} & \textbf{64.4} & \textbf{61.8} \\
        $\Delta$ (\textsc{Merlin}–MM) &
           $\snum{+}{2.0}$ & $\snum{+}{7.7}$ & $\snum{+}{5.9}$ & $\snum{+}{7.3}$ & $\snum{+}{8.7}$ &
           $\snum{+}{7.1}$ & $\snum{+}{5.0}$ & $\snum{+}{4.5}$ & $\snum{+}{4.4}$ & $\snum{+}{4.1}$ &
           $\mathbf{\snum{+}{5.2}}$ & $\mathbf{\snum{+}{5.9}}$ & $\mathbf{\snum{+}{5.6}}$ \\
        \midrule
        \multicolumn{14}{l}{\textit{Key Multilingual Approaches–Gemma 2 9B \& NLLB‑600M‑distilled}}\\
        QAlign-LoRA & 69.9 & 72.0 & 75.5 & 77.9 & 77.1 & 82.3 & 80.8 & 80.7 & 81.8 & 84.6 & 72.5 & 80.7 & 78.3\\
        SLAM          & 60.5 & 70.2 & 68.8 & 74.7 & 74.5 & 74.5 & 74.7 & 73.6 & 75.7 & 72.9 & 66.5 & 74.4 & 72.0 \\
        MindMerger    & 69.1 & 73.1 & 76.8 & \textbf{79.3} & 78.3 &
                         \textbf{82.6} & 79.5 & 79.5 & \textbf{82.5} & 82.3 &
                         73.0 & 80.6 & 78.3 \\
        \textbf{\textsc{Merlin}}& \textbf{72.1} & \textbf{76.1} & \textbf{77.4} & 78.7 & 79.8 & 81.5 &
                         \textbf{81.8} & \textbf{79.8} & 81.8 & \textbf{83.3} &
                         \textbf{75.2} & \textbf{81.0} & \textbf{79.2} \\
        $\Delta$ (\textsc{Merlin}–MM) &
           $\snum{+}{3.0}$ & $\snum{+}{3.0}$ & $\snum{+}{0.6}$ & $\snum{-}{0.6}$ & $\snum{+}{1.5}$ &
           $\snum{-}{1.1}$ & $\snum{+}{2.3}$ & $\snum{+}{0.3}$ & $\snum{-}{0.7}$ & $\snum{+}{1.0}$ &
           $\mathbf{\snum{+}{2.2}}$ & $\mathbf{\snum{+}{0.4}}$ & $\mathbf{\snum{+}{0.9}}$ \\
        \bottomrule
    \end{tabular}}
    \caption{Performance on MGSM and MSVAMP.  
    Bn/Th/Sw are mid‑resource (MRL); the other seven languages are high‑resource (HRL).  
    Sub‑headings (e.g.\ “Metamath‑7B \& mT5‑XL”) specify the \emph{decoder base model} and the \emph{multilingual encoder}.  
    LangBridge, LayAlign, MindMerger, and \textsc{Merlin} use both components; SLAM and QAlign adopt only the listed decoder.  
    $\Delta$ rows show \textsc{Merlin}’s absolute gain over MindMerger ($\snum{+}{x}$ for improvements, $\snum{-}{x}$ for drops).}
    \label{tab:mgsm_msvamp_combined}
    \vspace*{-2em}
\end{table*}

\section{Methodology}

Given a frozen multilingual encoder and a lightweight connector into the LLM’s embedding space, \textsc{Merlin} improves cross-lingual reasoning through a two-stage process. The first stage learns to align encoder outputs with the LLM representation space, while the second introduces a small adaptation inside the decoder to internalize this alignment for task solving. The subsections below describe the objectives and data used in each stage.

\subsection{Stage I: Model Stacking}
\label{sec:mapping}

Stage I is designed to help the decoder learn to interpret encoder representations without modifying its original parameters. This is accomplished through a three-step curriculum that gradually improves cross-lingual alignment, laying the groundwork for the subsequent task specialization stage, where the model’s reasoning capabilities are further refined.

\paragraph{Problem Definition: }Let $q$ be a target-language sentence of length $\ell$.  
A pretrained multilingual encoder produces $X=\operatorname{Enc}(q)\in\mathbb{R}^{\ell\times
d_{\text{enc}}}$.  A two-layer perceptron
$f_{\sigma}:\mathbb{R}^{d_{\text{enc}}}\!\rightarrow\!\mathbb{R}^{d_{\text{llm}}}$, with parameters
$\sigma$, projects these states into the LLM space, $X_f=f_{\sigma}(X)$.

The model stacking stage is divided into three consecutive sub-stages that progressively specialize the connector while the encoder and the LLM remain frozen.

\paragraph{ (a) General bilingual mapping (Map.)}
An initial bridge is established with a generic bitext from $L_{\text{tgt}}$, a generic sentence in the target language, to English  \citep{huang2024mindmergerefficientboostingllm}.  The
decoder receives only the projected encoder states
\[
  M_{\sigma}(q)=\bigl[\langle\texttt{bos}\rangle;\,X_f;\,\langle\texttt{sep}\rangle\bigr],
\]
and the connector parameters are obtained from
\[
  \sigma^{(1)}=\arg\!\min_{\sigma}
  \mathbb{E}_{(\tilde{s},s^{\star})}\!
  \bigl[-\log p_{\text{LLM}}\!\bigl(s^{\star}\mid M_{\sigma}(\tilde{s})\bigr)\bigr],
\]
where $\tilde{s}$ is the target-language sentence and $s^{\star}$ its English reference.  The loss
encourages $M_{\sigma}(\tilde{s})$ to be a close representation of $s^{\star}$ in the LLM space.

\paragraph{(b) Question Alignment \citep{zhu2024questiontranslationtrainingbetter} (Align.)}
The connector is then refined with translated prompts
$(\tilde{q},q^{\star})$, where $\tilde{q}$ is a
question in $L_{\text{tgt}}$ and $q^{\star}$ is its English counterpart:
\[
  \sigma^{(2)}=\arg\!\min_{\sigma}
  \mathbb{E}_{(\tilde{q},q^{\star})}\!
  \bigl[-\log p_{\text{LLM}}\!\bigl(q^{\star}\mid M_{\sigma}(\tilde{q})\bigr)\bigr].
\]
This step encourages the representations of non-English prompts to align closely with the semantic space utilized by the LLM for English reasoning.

\paragraph{(c) Task-aware Augmentation (Aug.)}
Finally, translated question–answer pairs $(\tilde{q},a^{\star})$ inject task-specific supervision.
Here the decoder is allowed to see both the encoder projection and its own token embeddings
$T(q)=E_{\text{LLM}}[q]$:
\[
  (X_f,T)(q)=\bigl[\langle\texttt{bos}\rangle;\,X_f;\,\langle\texttt{sep}\rangle;\,T(q)\bigr].
\]
The objective becomes
\[
\resizebox{1.0\linewidth}{!}{$
  \sigma^{(3)}=\mathop{\arg\!\min}\limits_{\sigma}
  \mathbb{E}_{(\tilde{q},a^{\star})}\!
  \bigl[-\log p_{\text{LLM}}\!\bigl(a^{\star}\mid (X_f,T)(\tilde{q})\bigr)\bigr],
$}
\]
yielding the final connector $\sigma^{\ast}=\sigma^{(3)}$.  


\subsection{Stage II: Task Specialization}
\label{sec:specialisation}

Stage II internalizes the cross-lingual signal inside the decoder; enhances the integration of multilingual representations within the decoder and plays a critical role in improving performance on complex reasoning tasks.  The connector
$\sigma^{\ast}$, the encoder and all pretrained LLM weights remain fixed; only rank-constrained DoRA  \citep{liu2024doraweightdecomposedlowrankadaptation}
matrices~$\theta$ inside the decoder are activated.
For every English task example $(q,a^{\star})$ the input takes the augmented form $(X_f,T)(q)$
defined above.  The adaptation parameters are obtained through
\[
\resizebox{1.0\linewidth}{!}{$
  \theta^{\ast}=\arg\!\min_{\theta}
  \mathbb{E}_{(q,a^{\star})}\!
  \bigl[-\log p_{\text{LLM}_{\theta}}\!\bigl(a^{\star}\mid (X_f,T)(q)\bigr)\bigr].
$}
\]
DoRA constrains learning to a negligible fraction of the total parameter count while letting the model
gradually absorb how to exploit $X_f$ during generation. 

\paragraph{Inference.}
Given a target-language query $q$, the encoder produces $X$, the connector adds the prefix
$X_f=f_{\sigma^{\ast}}(X)$, and the adapted decoder $\operatorname{LLM}_{\theta^{\ast}}$ generates
the answer in English from $(X_f,T)(q)$ \emph{zero-shot} with respect to $L_{\text{tgt}}$.

\subsection{Multi-stage instead of End-to-end design justification.}
Our multi-stage design is motivated by stability, data efficiency, and computational practicality. End-to-end training of the encoder $\to$ connector $\to$ decoder stack using limited LRL supervision tends to overfit and destabilize the coarse bilingual alignment learned from general bitext. Moreover, \textbf{Stage Ia is task-agnostic and can be reused} across math and NLI, whereas StageIb, Ic, and Stage II are lightweight task-specific steps; joint end-to-end training would entangle these and make reuse difficult.

\begin{table*}[ht]
    \centering
    \footnotesize
    \setlength{\tabcolsep}{4pt} 
    \resizebox{\textwidth}{!}{ 
        \begin{tabular}{@{}lccccccccccccccccccc@{}}
            \toprule
            \textbf{Model}  & \textbf{amh} & \textbf{ewe} & \textbf{hau} & \textbf{ibo} & \textbf{kin} & \textbf{lin} & \textbf{lug} & \textbf{orm} & \textbf{sna} & \textbf{sot} & \textbf{swa} & \textbf{twi} & \textbf{wol} & \textbf{xho} & \textbf{yor} & \textbf{zul} & \textbf{Avg} \\
            \midrule
            \multicolumn{10}{l}{\textit{closed models}} \\
            GPT-4o-mini & 31.6 & 6.0 & 56.0 & 33.6 & 48.0 & 25.6 & 29.2 & 39.2 & 44.8 & 36.8 & 70.8 & 15.6 & 7.6 & 32.4 & 45.6 & 43.6 & 35.4 \\
            GPT-4o & 57.6 & 8.8 & 64.8 & 57.6 & 60.4 & 51.2 & 51.6 & 61.2 & 58.4 & 60.8 & 78.8 & 31.2 & 28.0 & 52.4 & 62.0 & 57.2 & 52.6 \\
            \midrule
            \multicolumn{10}{l}{\textit{open models}} \\
            MetaMath-Mistral 7B & 1.2 & 2.4 & 2.0 & 2.0 & 2.4 & 3.6 & 4.4 & 2.0 & 1.2 & 2.4 & 10.4 & 2.8 & 2.8 & 1.6 & 1.6 & 1.6 & 2.8 \\
            Gemma 2 9B & 26.4 & 4.8 & 35.2 & 11.6 & 22.4 & 11.6 & 17.6 & 9.6 & 26.0 & 22.0 & 61.6 & 9.2 & 3.6 & 20.8 & 13.6 & 21.2 & 19.8 \\
            \midrule
            \multicolumn{10}{l}{\textit{Key multilingual approaches-MetaMath Mistral 7B \& AfriTeVa }} \\
            LangBridge & 31.6 & 2.0 & 41.2 & 14.0 & 28.4 & 4.0 & 11.2 & 17.2 & 30.0 & 29.6 & 54.4 & 2.8 & 0.8 & 18.4 & 21.2 & 27.2 & 20.9 \\   
            QAlign-LoRA & 4.4 & 3.6 & 3.2 & 1.6 & 5.6 & 7.6 & 3.2 & 4.8 & 5.6 & 4.8 & 18.4 & 2.4 & 3.6 & 4.0 & 4.4 & 4.8 & 5.1\\
            SLAM & 1.6 & 2.0 & 1.6 & 1.6 & 1.2 & 2.0 & 2.4 & 1.6 & 1.2 & 4.0 & 2.4 & 4.4 & 2.0 & 2.8 & 2.4 & 1.6 & 2.2 \\
            LayAlign & 30.0 & 6.4 & 36.4 & 27.2 & 32.0 & 16.0 & 21.6 & 26.8 & 31.6 & 30.0 & 42.8 & 4.0 & 6.0 & 23.6 & 30.4 & 29.6 & 24.6 \\
            MindMerger & 34.4 & 12.0 & 54.4 & 33.6 & 43.2 & 22.0 & 30.0 & 28.8 & 40.4 & 42.8 & \textbf{60.0} & \textbf{8.4} & 7.2 & 32.0 & 41.2 & 40.8 & 33.2 \\
            \textsc{Merlin} & \textbf{45.6} & \textbf{12.8} & \textbf{60.8} & \textbf{37.6} & \textbf{48.8} & \textbf{26.4} & \textbf{33.6} & \textbf{39.6} & \textbf{44.0} & \textbf{44.4} & 58.8 & 7.6 & \textbf{9.2} & \textbf{38.0} & \textbf{43.2} & \textbf{41.2} & \textbf{37.0} \\
            $\Delta$ (\textsc{Merlin}--MM) &
$\snum{+}{11.2}$ & $\snum{+}{0.8}$ & $\snum{+}{6.4}$ & $\snum{+}{4.0}$ & $\snum{+}{5.6}$ &
$\snum{+}{4.4}$ & $\snum{+}{3.6}$ & $\snum{+}{10.8}$ & $\snum{+}{3.6}$ & $\snum{+}{1.6}$ &
$\snum{-}{1.2}$ & $\snum{-}{0.8}$ & $\snum{+}{2.0}$ & $\snum{+}{6.0}$ & $\snum{+}{2.0}$ & $\snum{+}{0.4}$ &
$\mathbf{\snum{+}{3.8}}$ \\
            \midrule
            \multicolumn{10}{l}{\textit{Key multilingual approaches-Gemma2 9B \& NLLB-600M distilled}} \\
            QAlign-LoRA & 45.2 & 9.6 & 44.8 & 29.6 & 30.8 & 19.2 & 21.6 & 14.8 & 31.2 & 30.4 & 65.6 & 12.4 & 6.4 & 26.8 & 26.0 & 30.4 & 33.7 \\
            SLAM & 51.6 & 26.0 & 43.2 & 33.6 & 41.6 & 37.2 & 33.6 & \textbf{45.6} & 40.0 & 33.2 & 57.6 & 21.2 & 18.8 & 30.4 & 41.2 & 34.8 & 39.0 \\
            MindMerger & 45.2 & 32.4 & 44.8 & 34.4 & 43.2 & 38.4 & 36.4 & 40.0 & 44.8 & 36.8 & 53.2 & 24.4 & 21.2 & 33.6 & 34.8 & 40.4 & 37.7 \\
            \textsc{Merlin} & \textbf{59.2} & \textbf{41.2} & \textbf{60.8} & \textbf{52.0} & \textbf{52.0} & \textbf{49.6} &
      \textbf{46.8} & 32.0 & \textbf{58.8} & \textbf{56.0} & \textbf{76.8} & \textbf{37.6} & \textbf{26.0} & \textbf{51.2} & \textbf{54.0} & \textbf{55.6} & \textbf{50.6} \\
            $\Delta$ (\textsc{Merlin}--MM) &
$\snum{+}{14.0}$ & $\snum{+}{8.8}$ & $\snum{+}{16.0}$ & $\snum{+}{17.6}$ & $\snum{+}{8.8}$ &
$\snum{+}{11.2}$ & $\snum{+}{10.4}$ & $\snum{-}{8.0}$ & $\snum{+}{14.0}$ & $\snum{+}{19.2}$ &
$\snum{+}{23.6}$ & $\snum{+}{13.2}$ & $\snum{+}{4.8}$ & $\snum{+}{17.6}$ & $\snum{+}{19.2}$ & $\snum{+}{15.2}$ &
$\mathbf{\snum{+}{12.9}}$ \\
            \bottomrule
        \end{tabular}
    }
    \vspace*{-2mm}
    \caption{\textbf{AfriMGSM performance across various LLMs and key multilingual approaches}. We report Exact Match scores averaged over the displayed languages. We use the NLLB-600M distilled model as the encoder, and Gemma2-9b-it for LayAlign, MindMerger, \textsc{Merlin}, and as the base model for QAlign. We use AfriTeVa-v2-Large as the encoder of LangBridge. Note that the released LangBridge source code does not accept NLLB architecture.} 
    \label{tab:afirmgsm-all}
\end{table*}

\begin{table*}[ht]
    \centering
    \footnotesize
    \setlength{\tabcolsep}{4pt} 
    \resizebox{\textwidth}{!}{ 
        \begin{tabular}{@{}lccccccccccccccccccc@{}}
            \toprule
            \textbf{Model} & \textbf{eng} & \textbf{fra} & \textbf{amh} & \textbf{ewe} & \textbf{hau} & \textbf{ibo} & \textbf{kin} & \textbf{lin} & \textbf{lug} & \textbf{orm} & \textbf{sna} & \textbf{sot} & \textbf{swa} & \textbf{twi} & \textbf{wol} & \textbf{xho} & \textbf{yor} & \textbf{zul} & \textbf{Avg} \\
            \midrule
            \multicolumn{10}{l}{\textit{closed models}} \\
            GPT-4o-mini & 86.2 & 80.3 & 52.2 & 37.2 & 64.8 & 57.2 & 56.8 & 31.5 & 51.7 & 61.3 & 63.0 & 56.2 & 63.7 & 47.5 & 43.2 & 64.5 & 54.5 & 62.0 & 54.2 \\
            GPT-4o & 89.2 & 82.3 & 71.8 & 45.0 & \textbf{75.2} & 68.2 & 68.0 & 32.7 & 69.8 & \textbf{71.2} & 71.3 & 71.8 & 71.5 & 55.8 & 52.7 & 72.0 & 64.5 & 67.5 & 64.3 \\
            \midrule
            \multicolumn{10}{l}{\textit{open models}} \\
            Aya-101 & 67.0 & 59.7 & 64.2 & 43.2 & 57.0 & 55.5 & 54.3 & 33.5 & 51.7 & 51.5 & 55.7 & 52.2 & 56.5 & 47.0 & 36.7 & 55.2 & 54.5 & 55.3 & 51.5 \\
            Gemma 2 9B & 55.3 & 50.7 & 43.2 & 35.3 & 47.0 & 40.7 & 40.2 & 32.8 & 38.8 & 37.8 & 42.3 & 40.2 & 46.3 & 37.2 & 35.2 & 42.5 & 41.3 & 44.5 & 40.3 \\
            \midrule
            \multicolumn{10}{l}{\textit{Key Multilingual Approaches-Gemma2 \& NLLB-600m Distilled}} \\
            QAlign-LoRA  & 81.0 & 74.7 & 51.3 & 35.7 & 55.8 & 49.6 & 43.3 & 34.3 & 41.5 & 39.7 & 48.5 & 48 & 60.7 & 43.8 & 35.0 & 51.5 & 47.5 & 49.3 & 49.5\\ 
            SLAM & 67.0 & 61.8 & 48.2 & 42.7 & 47.2 & 51.3 & 41.7 & 32.2 & 47.5 & 47.7 & 46.5 & 47.2 & 52.8 & 38.8 & 35.7 & 49.8 & 52.2 & 48.3 & 47.7 \\ 
            MindMerger & 80.8 & 76.3 & 69.2 & \textbf{60.5} & 70.8 & 68.5 & 66.8 & 34.5 & \textbf{70.3} & 54.7 & \textbf{75.2} & 71.2 & 71.3 & 63.7 & 59.5 & \textbf{73.8} & 68.8 & \textbf{74.3} & 67.1 \\
            \textsc{Merlin}  & \textbf{90.8} & \textbf{86.7} & \textbf{74.8} & 60.0 & 73.3 & \textbf{71.7} & \textbf{68.8} & \textbf{34.8} & 69.2 & 59.3 & 71.3 & \textbf{72.0} & \textbf{72.2} & \textbf{65.5} & \textbf{62.2} & 73.7 & \textbf{69.2} & 70.2 & \textbf{68.7} \\
            \bottomrule
        \end{tabular}
    }
    \caption{\textbf{AfriXNLI performance on various LLMs, MindMerger, and \textsc{Merlin}}. Metric is Accuracy. Average computed on only African languages. NLLB-600M distilled is used as the encoder.}
    \label{tab:afrixnli}
    \vspace*{-1.5em}
\end{table*}


\section{Experimental Setup}

\subsection{LLMs evaluated} 
We evaluate two closed models(GPT-4o-mini and GPT-4o) 
and four open LLMs, including Aya-101~\citep{ustun2024ayamodelinstructionfinetuned}, MetaMath 7B, MetaMath-Mistral 7B~\citep{yu2023metamath}, and Gemma 2 9B~\citep{gemmateam2024gemma2improvingopen}. Aya-101 is only evaluated for the AfriXNLI as it is one of the strongest open model baselines for low-resource languages. 

Our choice of MetaMath 7B/Mistral and Gemma 2 9B is based on their strong English mathematical reasoning abilities, which is crucial because \textsc{Merlin}’s goal is to transfer existing reasoning competence from English to target languages. Many general-purpose models (e.g., LLaMA 3.1 \cite{grattafiori2024llama3herdmodels}) underperform on GSM-style math \cite{adelani2024irokobenchnewbenchmarkafrican}, leaving little reasoning skill to transfer. Overall, \textsc{Merlin} is backbone-agnostic and can be applied to other LLMs in principle.

\subsection{Baselines Method Compared}

We evaluate our method against five strong baselines: three model merging approaches---LangBridge~\protect\citep{yoon-etal-2024-langbridge}, MindMerger~\protect\citep{huang2024mindmergerefficientboostingllm},  LayAlign~\protect\citep{ruan2025layalignenhancingmultilingualreasoning} and two methods leveraging in-domain question translation data, QAlign-LoRA~\citep{zhu2024questiontranslationtrainingbetter} and SLAM~\citep{fan2025slamefficientmultilingualreasoning}. \textbf{LangBridge} injects a frozen multilingual encoder before a decoder-only LLM, replacing token embeddings with linearly projected encoder states ($\approx$3M parameters) learned from 100k English instructions, and inference cost equals that of the LLM plus the multilingual encoder. \textbf{QAlign-LoRA} first aligns multilingual prompts via question translation and then fine-tunes lightweight LoRA adapters (<1\% parameters) on English tasks. \textbf{SLAM} fine-tunes the six lowest feed-forward layers ($\approx$7\% parameters) identified as language-sensitive on mixed English and translated data, freezing higher layers and attention weights. \textbf{MindMerger} connects a frozen multilingual encoder to the LLM through a two-layer MLP head ($\approx$9M parameters), trained first on translation pairs with English references, then on translated QA pairs while keeping all base weights fixed. \textbf{LayAlign} integrates encoder features into multiple LLM layers via a trainable layer-wise aligner, trained in two stages using translation and task data while keeping both encoder and LLM frozen.

For the encoder component in multilingual‐integration approaches, we couple both \textsc{MindMerger} and \textsc{Merlin} with the distilled 600M‐parameter \textsc{NLLB} 
encoder~\citep{nllbteam2022languageleftbehindscaling}.  For \textsc{LangBridge}, we follow its original setup: AfriTeva-v2-Large~\citep{oladipo-etal-2023-better} is used on \texttt{AfriMGSM} and \texttt{AfriXNLI}, whereas mT5-XL is employed for \texttt{MGSM}, since the publicly released LangBridge implementation only supports encoders of the class \texttt{XXXEncoderModel} (e.g., \texttt{MT5EncoderModel}). 

\subsection{Languages and Tasks} 
Our evaluation spans four multilingual benchmarks. \textbf{AfriMGSM}~\citep{adelani2024irokobenchnewbenchmarkafrican} extends \textsc{MGSM} with grade-school mathematics problems translated into 16 African languages, providing fine-grained coverage of more low-resource settings. \textbf{AfriXNLI}~\citep{adelani2024irokobenchnewbenchmarkafrican} offers natural-language inference data in the same 16 languages by translating ten domain splits from the original \textsc{XNLI}. The canonical \textbf{MGSM} corpus~\citep{shi2022languagemodelsmultilingualchainofthought} supplies math QA instances across eight mid-resource languages. Finally, \textbf{MSVAMP}~\citep{chen2024breakinglanguagebarriersmultilingual} introduces multilingual variants of SVAMP with 1000 algebra-word problems in eleven languages, testing cross-lingual reasoning beyond GSM-style arithmetic.

\subsection{Experimental Settings}

\paragraph{Training Datasets}

\textbf{Sub-stage Ia (Mapping).}  
For every language we sample 9\,000 general-domain English–target sentence pairs from the \textsc{NLLB} training corpus \citep{nllbteam2022languageleftbehindscaling}.  These bitext pairs supply a coarse bilingual anchor for the connector.

\paragraph{Sub-stage Ib (Question Alignment).}  
To align non-English prompts with English semantics, we translate 20\,000 English questions with \texttt{NLLB}\,200-3.3B and sample 3\,000 translated pairs per language. For the mathematics track (\texttt{AfriMGSM} and \texttt{MGSM}) the source questions come from \texttt{MetaMathQA} \citep{yu2024metamathbootstrapmathematicalquestions}. For the NLI track (\texttt{AfriXNLI}) we instead sample premise–hypothesis pairs from translated \texttt{MultiNLI} \citep{williams2018broadcoveragechallengecorpussentence}, concatenate them as the “question”, translate, and again retain 3\,000 pairs. 

\paragraph{Sub-stage Ic (Augmentation).}  
For each target language we extract a distinct set of 3\,000 question–answer pairs from the training split of the task corpus: translated \texttt{MetaMathQA} for the math benchmarks (\texttt{AfriMGSM}, \texttt{MGSM}, and \texttt{MSVAMP}) and translated \texttt{MultiNLI} for the NLI benchmark (\texttt{AfriXNLI}). 

\paragraph{Task-Specialization Stage.}  
We fine-tune the LLM (via PEFT) on a second sample of 3\,000 training instances per language drawn from the same translated sources as above. Using a separate set of examples helps reduce the risk of overlap with Sub-stage~Ic and ensures there is no data leakage into the evaluation sets.\footnote{The detailed training configuration is provided on Appendix}

\paragraph{Alignment of training data across methods.} \textsc{Merlin} and all baselines are trained on the same underlying supervision. For math tasks, all methods (\textsc{MindMerger}, \textsc{QAlign}, \textsc{SLAM}, \textsc{Merlin}) use the same question-level translations derived from MetaMathQA; for AfriXNLI, all methods use premise–hypothesis pairs derived from MultiNLI. The only difference lies in how each method utilizes this data (e.g., which module is trained at which stage).

\begin{table*}[t]
  \centering
  \begin{adjustbox}{width=\textwidth}
  \begin{tabular}{c c c c|*{18}{r}|r}
    \toprule
    \textbf{Map} & \textbf{Align} & \textbf{Aug} & \textbf{Spe} &
    \textbf{eng} & \textbf{fra} & \textbf{amh} & \textbf{ewe} &
    \textbf{hau} & \textbf{ibo} & \textbf{kin} & \textbf{lin} &
    \textbf{lug} & \textbf{orm} & \textbf{sna} & \textbf{sot} &
    \textbf{swa} & \textbf{twi} & \textbf{wol} & \textbf{xho} &
    \textbf{yor} & \textbf{zul} & \textbf{Avg} \\
    \midrule
    \cmark & \xmark & \cmark & \xmark &
58.0 & 50.8 & 45.2 & 32.4 & 44.8 & 34.4 & 43.2 & 38.4 &
36.4 & \textbf{40.0} & 44.8 & 36.8 & 53.2 & 24.4 & 21.2 & 33.6 &
34.8 & 40.4 & 39.60 \\
    \xmark & \cmark & \cmark & \cmark &
      \textbf{82.8} & \textbf{78.0} & 52.8 & 34.0 & 60.4 & 50.4 & 52.0 & 47.2 &
      \textbf{46.8} & 24.0 & 52.8 & 53.2 & 73.6 & 34.8 & 23.6 & 43.2 & 53.6 & 52.4 & 50.87 \\
    \cmark & \xmark & \cmark & \cmark &
      86.0 & 81.2 & \textbf{59.6} & 37.2 & 59.6 & 50.8 & \textbf{52.8} & 45.2 &
      44.4 & 30.4 & 54.0 & 53.6 & 73.2 & 36.8 & 23.6 & 48.8 & 53.6 & 54.0 & 52.49 \\
    \cmark & \cmark & \xmark & \cmark &
      77.2 & 64.0 & 26.8 & 16.0 & 30.8 & 20.0 & 28.0 & 19.2 &
      20.3 & 13.9 & 24.7 & 24.5 & 33.5 & 16.8 & 10.8 & 22.3 & 24.5 & 24.7 & 27.67 \\
    \cmark & \cmark & \cmark & \xmark &
      80.0 & 73.2 & 57.2 & 35.6 & 56.8 & 48.0 & 57.2 & 44.4 & 45.2 & 30.4 & 57.6 & 55.2 & 74.0 & 34.0 & 24.0 & 48.4 & 45.2 & 50.4 & 50.93 \\
    \cmark & \cmark & \cmark & \cmark &
    82.0 & 74.0 & 59.2 & \textbf{41.2} & \textbf{60.8} & \textbf{52.0} & 52.0 & \textbf{49.6} &
      \textbf{46.8} & 32.0 & \textbf{58.8} & \textbf{56.0} & \textbf{76.8} & \textbf{37.6} & \textbf{26.0} & \textbf{51.2} & \textbf{54.0} & \textbf{55.6} & \textbf{53.64} \\
    \bottomrule
  \end{tabular}
  \end{adjustbox}
  \caption{Ablation study: each row removes one stage (\xmark) while keeping the others (\cmark).  
Map = General Bilingual Mapping; Align = Question Alignment;  
Aug = Task-aware Augmentation; Spe = Task Specialization Stage. The first row corresponds to the MindMerger experiment setting.}
  \label{tab:full_ablation}
  \vspace*{-5mm}
\end{table*}

\section{Experimental Results}

We assessed \textsc{Merlin} on both math reasoning and NLI
benchmarks, comparing it against open LLMs, recent multilingual
integration methods, and the strongest available closed models. For math reasoning, we use two datasets: MGSM and MSVAMP. To assess performance on low-resource African languages across both tasks, we employ AfriMGSM and AfriXNLI.
All results are reported as exact-match or accuracy; higher is better.

\paragraph{\textsc{Merlin} achieved SOTA on standardized math reasoning benchmarks}
Table~\ref{tab:mgsm_msvamp_combined} presents results on MGSM and its
algebraic variant MSVAMP.  
Using the identical Gemma 2 9B backbone, \textsc{Merlin} achieves
\(76.2\,\%\) average accuracy on MGSM, exceeding
MindMerger by \(+2.8\) points and SLAM by \(+22.8\) points.%
\footnote{LangBridge and QAlign scores are omitted in
Table~\ref{tab:mgsm_msvamp_combined} because their publicly released
implementations do not support the full language set.}
Improvements are consistent across both mid-resource languages
(Bn/Th/Sw, \(+13.7\) over Gemma 2) and high-resource languages
(\(+7.7\) over MetaMath-Mistral).
On MSVAMP, \textsc{Merlin} retains its lead with \(79.2\,\%\) average accuracy,
outperforming MindMerger by \(+0.9\) points and exceeding the best open
baseline by more than \(15\) points.  These gains indicate that the two-stage
alignment not only transfers arithmetic skills but also generalises to
algebraic word problems.

\paragraph{\textsc{Merlin} achieved a wider performance gap in math reasoning for LRLs}
Table~\ref{tab:afirmgsm-all} focuses on sixteen
truly low-resource African languages.  
With the strong Gemma 2 9B+NLLB-600M configuration,
\textsc{Merlin} attains \(50.6\,\%\) average accuracy, surpassing MindMerger by
\(+12.9\) points, SLAM by \(+11.6\) points, and QAlign-LoRA by
\(+16.9\) points.  
Notably, \textsc{Merlin} outruns the closed GPT-4o-mini by \(+15.2\) points and
comes within \(2.0\) points of GPT-4o despite the latter’s much larger
parameter budget.
Ablation confirms (\autoref{tab:full_ablation}) that the
task-aware augmentation sub-stage is decisive, yielding gains of more
than ten points when added to the connector training.


\paragraph{Reasoning capabilities generalize to NLU tasks}
Table \ref{tab:afrixnli} reports accuracy on sixteen African
languages plus English and French on NLI task.  \textsc{Merlin} attains
\(68.7\,\%\) average accuracy on the African subset,
improving on MindMerger by \(+1.6\) points and on the base
Gemma 2 9B model by \(+28.4\) points, and marginally surpassing the
closed GPT‑4o (\(64.3\,\%\)).  The gains are most pronounced for
\emph{Amharic} (+5.6), \emph{Oromo} (+4.6), \emph{Hausa} (+2.5),
and \emph{Kinyarwanda} (+2.0), confirming the
value of the cross‑lingual mapping curriculum in truly low‑resource
settings.  The overall improvement across
thirteen of the sixteen African languages indicates that \textsc{Merlin}
successfully redistributes modelling capacity toward languages that
benefit the most from additional cross‑lingual supervision, while
maintaining competitive performance on those already well supported.




\begin{figure}[htp!]
    \centering
    \includegraphics[width=1\columnwidth]{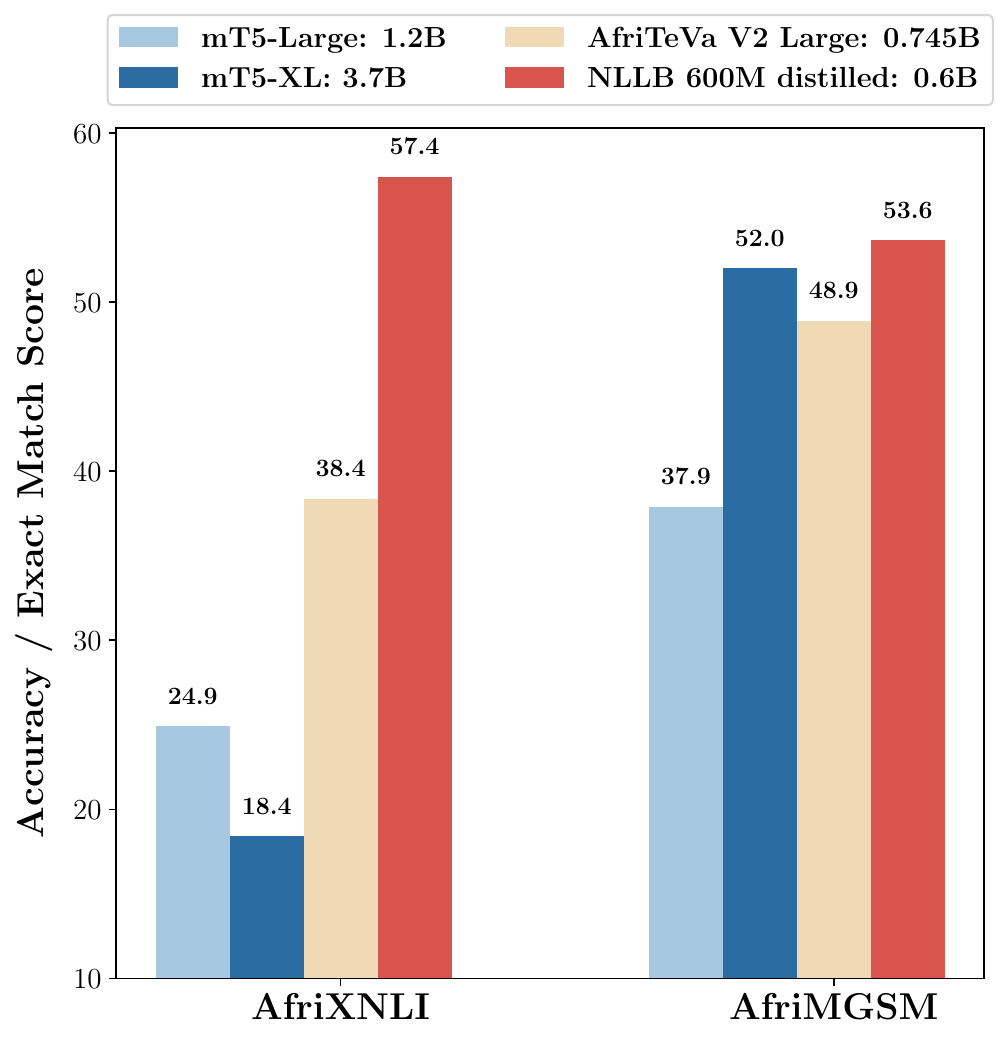}
    \vspace{-2em}
    \caption{Comparison of \textsc{Merlin} performance across five different multilingual encoders and Gemma 2 9B LLM.}
    \label{fig:barplot}
    \vspace{-1.5em}
\end{figure}

\section{Analysis of Language Alignment and Effects on Multilingualism}



\begin{figure*}[htp!]
    \centering
    \includegraphics[width=\linewidth]{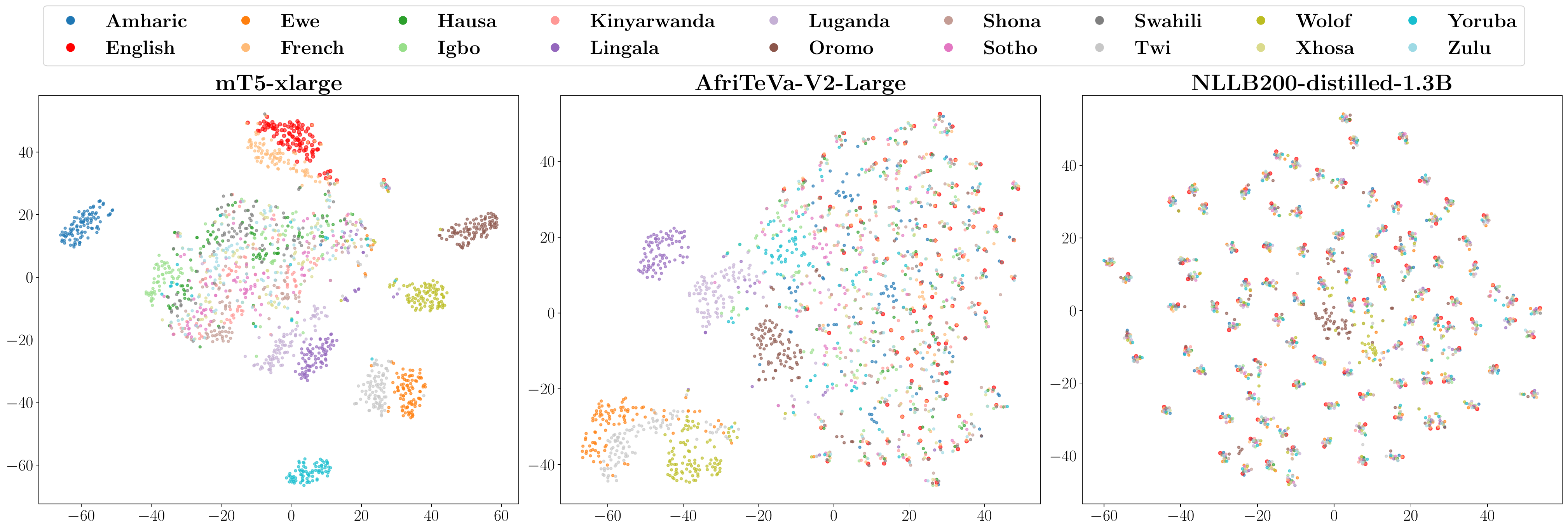}
    \vspace*{-2em}
    \caption{T-SNE visualizations of sentence embeddings for mT5-xl, AfriTeVa-V2-Large, and NLLB200-distilled-1.3B, showing how LRL data impacts cross-lingual alignment. More data reduces clustering, bringing low-resource languages closer to English, enhancing transfer learning.}
    \label{fig:tsne_visualization}
    \vspace*{-2mm}
\end{figure*}



\paragraph{Effectiveness in Low and Mid-resource reasoning Settings} 
Across all three evaluation results (Table \ref{tab:mgsm_msvamp_combined}, \ref{tab:afirmgsm-all}), \textsc{Merlin} delivers its greatest improvements in the settings with lower resource settings. On \textsc{MGSM}, the model improves over \textsc{MindMerger} by \textbf{+6.5}~accuracy points for the mid-resource group (Bn, Th, Sw), while the corresponding gain for the seven high-resource languages is only \textbf{+1.3}.  
The same pattern holds for \textsc{MSVAMP} (\textbf{+2.2} vs.\ \textbf{+0.4}).  
The advantage widens on the 16-language \textsc{AfriMGSM} benchmark, where \textsc{Merlin} averages a \textbf{+12.9}-point delta and delivers double-digit improvements in ten languages.  
We attribute the larger mid and low-resource gains to the combination of \emph{Question Alignment} (Sub-stage~Ib) and the PEFT–based \emph{Task Specialization} of Stage~II. 
The former links each non-English question to an English prompt of identical structure, improving cross-lingual grounding, while the latter tunes a small set of low-rank weights to consolidate this grounding inside the decoder without disturbing the frozen backbone.  
Evidence comes from the ablations in Table~\ref{tab:full_ablation}: disabling Ib (\textsc{Map}\;\cmark\;\textsc{Align}\;\xmark\;\textsc{Aug}\;\cmark\;\textsc{Spe}\;\cmark) or Stage II (\textsc{Map}\;\cmark\;\textsc{Align}\;\cmark\;\textsc{Aug}\;\cmark\;\textsc{Spe}\;\xmark) reduces the overall score from 53.6 to 50.9 and 52.5, respectively, with the steepest drops concentrated in the low-resource languages. We further compare \textsc{Merlin} with MindMerger under identical data volumes to show that \textsc{Merlin}’s performance gains arise from its three-stage training curriculum rather than increased data (Appendix~\ref{sec:merlin_isolate_curriculum_and_data}).

\paragraph{Encoders rich in LRL data sharpen Cross-lingual alignment}

A key factor in multilingual model merging is how different encoder choices affect the representation of LRLs in the mapping layer. To isolate this effect, we kept the \textsc{Merlin} architecture unchanged across all experiments and varied only the multilingual encoder. As shown in Figure \ref{fig:tsne_visualization}, when the proportion of LRL text is small, non-English data---especially from LRLs---forms distinct clusters far from English in the LLM’s original embedding space. This separation prevents the model from sharing or transferring internal reasoning paths learned predominantly through English data. Conversely, incorporating more low-resource data enables the mapping layer to align different language embeddings, including those from LRLs, more closely with English. By bringing LRLs into the same representational space as English, the LLM can more effectively leverage its inherent reasoning capabilities for LRL tasks, rather than leaving them isolated at the margins of the embedding space. Consequently, increasing low-resource language datasets during encoder training promotes cross-lingual knowledge transfer and enhances representation consistency across languages.

\paragraph{Middle Decoder layers reach peak Cross-lingual alignment}

\begin{figure}[htp!]
    \centering
    \includegraphics[width=1\columnwidth]{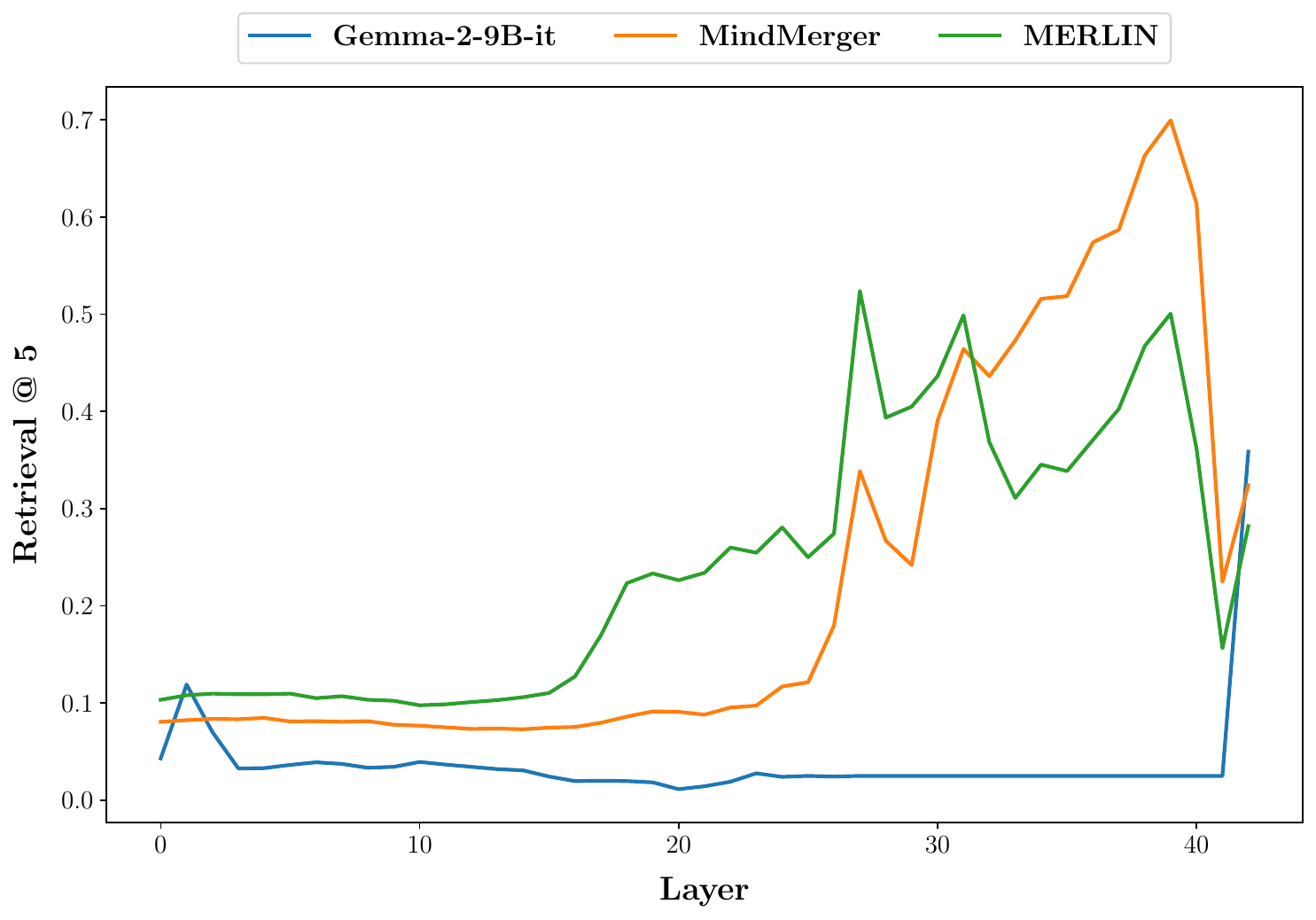}
    \vspace*{-2em}
    \caption{\textbf{Layer-wise cross-lingual retrieval@$5$.}  
           Scores are averaged over English and 16 African languages.  
           \textsc{Merlin} achieves the highest mid-layer alignment, whereas MindMerger peaks only in the final decoder blocks.}
    \label{fig:retrieval5}
    \vspace*{-1.5em}
\end{figure}

Figure~\ref{fig:retrieval5} reports sentence-level retrieval@5 for every decoder layer of \textsc{Gemma2-9B-It}, \textsc{MindMerger}, and \textsc{Merlin}.  

Retrieval@5 is computed by taking each English sentence in the FLORES-200 dev set as a query, ranking all candidate sentences in the paired target language by the cosine similarity of their layer-wise embeddings, and then checking whether the gold translation appears among the top-5 nearest neighbours. The score---averaged over 200 sentence pairs and 15 African languages---therefore reflects the probability that a representation keeps the correct cross-lingual match within a five-item shortlist, providing a direct proxy for how language-agnostic the layer’s embedding space is.

From layers~18--28 \textsc{Merlin} reaches its peak sentence--level retrieval@5 (\raisebox{.15em}{\texttildelow}0.30).  
The rise coincides with the PEFT fine–tuning in Stage~II, which adjusts a small set of low-rank weights on top of the frozen LLM while keeping the connector $f_{\sigma^{\star}}$ fixed.  
This adaptation strengthens the multilingual abstractions that naturally emerge in the decoder’s middle blocks---the region reported to be the most language-agnostic once parallel supervision is available \citep{liu2025middlelayerrepresentationalignmentcrosslingual}.  
As a result, non-English queries are already embedded in the English reasoning sub-space before the later, generation-oriented layers are reached, providing a solid cross-lingual scaffold for downstream mathematical reasoning tasks such as \textsc{AfriMGSM}.


\paragraph{Isolating the Effects of Curriculum Structure and Data Volume}
\label{sec:merlin_isolate_curriculum_and_data}

To disentangle curriculum effects from data volume, we compare \textsc{Merlin} and MindMerger under an identical supervision budget of 12k examples per language. As shown in Table~\ref{tab:merlin_isolate_curriculum_and_data}, \textsc{Merlin} still consistently outperforms MindMerger on both MGSM (78.3 vs.\ 73.4) and MSVAMP (80.4 vs.\ 78.3), despite using no additional data.

Since both methods use the same total supervision, the gains cannot be attributed to data quantity. Instead, the key difference lies in how supervision is organized: MindMerger concentrates most data in a single bilingual mapping stage, whereas \textsc{Merlin} distributes supervision across multiple curriculum stages that progressively transition from general alignment to task-specific reasoning. The consistent improvements across mid- and high-resource languages indicate that this staged curriculum yields more effective cross-lingual and task-aligned representations, validating curriculum structure as the primary source of \textsc{Merlin}'s gains.

 \begin{table*}[!ht]
    \centering
    \tiny                      
    \setlength{\tabcolsep}{4pt}
    \renewcommand{\arraystretch}{0.8}
    \resizebox{\textwidth}{!}{
    \begin{tabular}{@{}lcccccccccc|ccc@{}}
        \toprule
        \textbf{Model} & \textbf{Bn} & \textbf{Th} & \textbf{Sw} & \textbf{Ja} &
        \textbf{Zh} & \textbf{De} & \textbf{Fr} & \textbf{Ru} &
        \textbf{Es} & \textbf{En} & \textbf{Mrl.} & \textbf{Hrl.} & \textbf{Avg} \\
        \midrule
        \multicolumn{14}{l}{\textbf{MGSM}}\\
        MindMerger  & 66.8 & 72.8 & 69.2 & 66.8 & 72.8 & 78.0 & 75.2 & 73.6 & 76.4 & 82.4 & 69.6 & 75.0 & 73.4 \\
        MERLIN (1k / stage) & \textbf{76.4} & \textbf{77.6} & \textbf{74.4} & \textbf{71.6} & \textbf{76.0} & \textbf{78.4} & \textbf{77.6} & \textbf{81.6} & \textbf{83.6} & \textbf{85.6} & \textbf{76.1} & \textbf{79.2} & \textbf{78.3} \\
        \midrule
        \multicolumn{14}{l}{\textbf{MSVAMP}}\\
        MindMerger  & 69.1 & 73.1 & 76.8 & 79.3 & 78.3 & 82.6 & 79.5 & 79.5 & 82.5 & 82.3 & 73.0 & 80.6 & 78.3 \\
        MERLIN (1k / stage) & \textbf{73.0} & \textbf{75.7} & \textbf{79.4} & \textbf{79.7} & \textbf{82.2} & \textbf{82.9} & \textbf{82.7} & \textbf{81.6} & \textbf{82.8} & \textbf{84.4} & \textbf{76.0} & \textbf{82.3} & \textbf{80.4} \\
        \bottomrule
    \end{tabular}}
    \caption{Full per-language comparison between MindMerger and \textsc{Merlin} using the same total amount of supervision (12k pairs) per language on MGSM and MSVAMP. MindMerger uses 9k bilingual pairs (mapping) + 3k QA (task-aware), whereas \textsc{Merlin} (1k / stage) uses 9k bilingual pairs + 1k for each of Stage Ib, Stage Ic, and Stage II. Bn/Th/Sw are mid-resource (MRL); Ja/Zh/De/Fr/Ru/Es/En are high-resource (HRL).}
    \label{tab:merlin_isolate_curriculum_and_data}
\end{table*}

\paragraph{Disentanglement between Understanding and Reasoning abilities of LLM}
Since MGSM and AfriMGSM are direct translations of GSM8K, all models generate the final answer in English. This means the underlying reasoning task is identical across languages, so any performance gap cannot stem from limitations in the LLM’s reasoning ability. Instead, it directly reflects the quality of cross-lingual alignment. The above results strongly support this interpretation: English accuracy remains stable across all methods, while \textsc{Merlin} delivers the largest gains in low- and mid-resource languages. \textsc{Merlin} also improves substantially on AfriXNLI, a setting focused on sentence-level understanding rather than multi-step reasoning, further indicating that the curriculum strengthens cross-lingual representations rather than altering the LLM’s reasoning behavior. Our layer-wise retrieval and encoder-choice analyses reinforce this conclusion by showing clearer alignment in the decoder’s intermediate layers.

\section{Related Work}

\paragraph{LLMs in LRLs} LRLs often suffer from limited annotated data and high-quality machine translation resources, creating significant challenges for NLP models \citep{magueresse2020lowresourcelanguagesreviewpast,lee_pre-trained_2022,app13158566,deshpande2024chainoftranslationpromptingcotrnovel, qin2024multilinguallargelanguagemodel, tonja-etal-2024-nlp}. Moreover, these languages can exhibit complex morphological structures and orthographic variations \citep{ghosh2024morphologybasedinvestigationpositionalencodings, nzeyimana2024lowresourceneuralmachinetranslation, lopo2024constructingexpandinglowresourceunderrepresented, issaka2024ghanaiannlplandscapelook, lusito2023textnormalizationlowresourcelanguages}, posing further challenges. While multilingual models can handle many languages, they frequently underperform in truly low-resource settings and may even trail behind strong monolingual baselines \citep{yoon-etal-2024-langbridge, du2024chinesetinyllmpretraining, blevins-etal-2024-breaking}. Existing efforts to boost LRL performance focus primarily on dataset creation \citep{adelani-etal-2021-masakhaner,muhammad2022naijasentinigeriantwittersentiment, adelani-etal-2024-sib,yong2024lexcgengeneratingdataextremely} or adaptation of pretrained models \citep{alabi-etal-2022-adapting, wu2024adaptinglargelanguagemodels, csaki2023efficientlyadaptingpretrainedlanguage,schmidt-etal-2024-self}, yet these strategies remain constrained by data scarcity and fail to fully address complex linguistic properties \citep{urbizu-etal-2023-scaling, wang2023challengingbenchmarklowresourcelearning}.


\paragraph{Multilingual Reasoning.} Recent efforts to enhance multilingual \emph{reasoning} in LLMs include carefully devised prompt engineering to facilitate multi-step inference or chain-of-thought reasoning \citep{huang-etal-2023-languages, qin-etal-2023-cross}, although these techniques often lose effectiveness with smaller open-source models \citep{touvron2023llama2openfoundation}. Another approach involves supervised fine-tuning on translated data \citep{chen2024breakinglanguagebarriersmultilingual, zhu2024questiontranslationtrainingbetter}, which infuses language or task-specific knowledge into the model. Building upon this, \citet{fan2025slamefficientmultilingualreasoning} extend the work of \citet{zhu2024questiontranslationtrainingbetter} by optimizing both the training parameters and the training stages focusing on language-specific neurons. Their improvements lead to higher average performance on multilingual mathematical reasoning benchmarks.
However, reliance on machine translation can introduce errors \citep{shi2022languagemodelsmultilingualchainofthought} and may overlook the LLM’s built-in multilingual capabilities. To circumvent these issues, MindMerger \citep{huang2024mindmergerefficientboostingllm} sidesteps autoregressive translation by directly integrating an external multilingual encoder, preserving the LLM’s internal parameters. 



\section{Conclusion}
\label{sec:conclusion}

This work introduced \textsc{Merlin}, a lightweight framework for strengthening multilingual mathematical reasoning and natural‑language inference in large language models.  A two‑stage curriculum first learns a compact connector that projects multilingual encoder states into the frozen LLM embedding space, then refines the decoder through DoRA‑based task specialization.  The approach preserves the original English reasoning capabilities while substantially improving performance in both mid‑ and low‑resource languages.

Our analysis shows that model merging enhances cross-lingual alignment across layers compared to using LLMs alone. In particular, our framework, \textsc{Merlin}, improves alignment within the model’s internal reasoning subspace, yielding stronger performance on reasoning tasks even under tightly matched supervision budgets. We also find that the choice of encoder is critical for low-resource languages: encoders with greater exposure to LRL data produce representations that align more closely with the LLM’s internal language (often English), thereby supplying richer cross-lingual signals for model merging.

Empirical evaluation on four benchmarks demonstrates that \textsc{Merlin} outperforms strong open‑source systems---including MindMerger, SLAM, and LangBridge---on MGSM, MSVAMP, AfriMGSM and AfriXNLI, and narrows the gap to closed models such as GPT‑4o.

\section{Limitations}
\label{sec:limitations}

\textsc{Merlin} relies on automatically translated corpora for all three training stages, assuming that these translations form a reliable semantic bridge between the target language and English. In practice, translation quality is uneven. Even with a strong multilingual model such as the distilled \textsc{NLLB} 600M encoder, we observe systematic errors for certain languages---most notably Oromo---where mistranslations propagate through the mapping layer and erode downstream accuracy. Because the present pipeline admits every machine-translated sentence without verification, noisy lexical choices and hallucinated numerals enter the training signal unchecked. Mitigating this weakness will require explicit data filtering.

A second limitation is the narrow scope of our evaluation. The experiments target two problem families---grade-school mathematics and natural-language inference---selected for their diagnostic value in cross-lingual reasoning. Tasks that demand other forms of structured reasoning, such as program synthesis, commonsense QA, or multi-document summarization, remain unexplored. 

Finally, our methodology trains a separate \textsc{Merlin} instance for each benchmark. This design clarifies the contribution of the mapping curriculum but precludes parameter sharing across tasks and prevents the model from exploiting potential synergies between, for example, symbolic arithmetic and commonsense inference. A unified multi-task variant would be more practical in deployment and could yield additional gains through cross-task transfer, although its optimization dynamics are not yet understood.





\section*{Acknowledgments}
This research was supported in part by the Natural Sciences and Engineering Research Council (NSERC) of Canada. David Adelani acknowledges the funding of IVADO and the Canada First Research Excellence Fund. We acknowledge the use of generative AI tools for improving proofreading and readability, no scientific content was generated by the tools.

\bibliography{custom}

\appendix











\section{Dataset-Size Ablation}
\label{app:size_ablation}

With the full sweep of dataset sizes in Table \ref{tab:merlin_size_ablation}, a clear elbow appears at \textbf{1 000} translated sentence pairs per language and per stage. At that level \textsc{Merlin} attains its highest averages—78.3 \% on MGSM and 80.4 \% on MSVAMP—while using only one-sixth of the data required by the largest setting. A modest increase to 1 500 pairs produces small, mixed fluctuations, and every larger tier either plateaus or declines, a pattern we attribute to domain mismatch and over-fitting of the lightweight connector once the easiest bilingual correspondences have been captured. Even the 500-pair condition is competitive (75.2 \% / 80.1 \%), suggesting that the first few hundred high-quality translations already establish a usable cross-lingual bridge. In terms of compute, the 1 000-pair configuration is exceptionally light: all three connector stages plus DoRA adaptation complete in just under one hour on a single A100-80 GB GPU (batch size = 1), confirming that \textsc{Merlin} can be trained end-to-end in resource-constrained settings without sacrificing downstream accuracy.

\begin{table*}[!ht]
    \centering
    \tiny                      
    \setlength{\tabcolsep}{4pt}
    \renewcommand{\arraystretch}{0.8}
    \resizebox{\textwidth}{!}{
    \begin{tabular}{@{}lcccccccccc|ccc@{}}
        \toprule
        \textbf{Model / Size} & \textbf{Bn} & \textbf{Th} & \textbf{Sw} & \textbf{Ja} &
        \textbf{Zh} & \textbf{De} & \textbf{Fr} & \textbf{Ru} &
        \textbf{Es} & \textbf{En} & \textbf{Mrl.} & \textbf{Hrl.} & \textbf{Avg} \\
        \midrule
        \multicolumn{14}{l}{\textbf{MGSM}}\\
        500  & 72.8 & 73.6 & 71.6 & 72.4 & 73.2 & 74.8 & 75.6 & 79.6 & 76.8 & 82.0 & 72.7 & 76.3 & 75.2 \\
        1000 & \textbf{76.4} & 77.6 & 74.4 & 71.6 & \textbf{76.0} & \textbf{78.4} & 77.6 & \textbf{81.6} & 83.6 & \textbf{85.6} & 76.1 & \textbf{79.2} & \textbf{78.3} \\
        1500  & 75.6 & 74.4 & 71.6 & 72.8 & 76.4 & 74.4 & 75.6 & 78.8 & \textbf{85.2} & 84.8 & 73.9 & 78.3 & 77.0 \\
        3000& 74.8 & 76.0 & 77.6 & 71.6 & 72.8 & 73.2 & 74.8 & 78.4 & 77.6 & \textbf{85.6} & 76.1 & 76.3 & 76.2 \\
        4500 & 74.8 & 77.6 & 72.8 & \textbf{74.0} & 74.4 & 75.6 & 75.6 & 78.4 & 82.8 & 84.0 & 75.1 & 77.8 & 77.0 \\
        9000   & 74.8 & 78.4 & 74.8 & \textbf{74.0} & 72.4 & 76.0 & 74.8 & 77.6 & 81.2 & 83.6 & 76.0 & 77.1 & 76.8 \\
        18000  & 76.0 & \textbf{79.6} & \textbf{79.2} & 73.2 & 72.8 & 76.8 & \textbf{79.2} & 78.0 & 82.8 & 83.6 & \textbf{78.3} & 78.1 & 78.1 \\
        \midrule
        \multicolumn{14}{l}{\textbf{MSVAMP}}\\
        500  & 72.6 & 75.3 & 77.8 & 80.7 & 81.2 & 82.8 & 82.5 & 81.0 & 83.1 & 83.5 & 75.2 & 82.1 & 80.1 \\
        1000 & \textbf{73.0} & 75.7 & \textbf{79.4} & 79.7 & \textbf{82.2} & \textbf{82.9} & \textbf{82.7} & 
        \textbf{81.6} & \textbf{82.8} & 84.4 & \textbf{76.0} & \textbf{82.3} & \textbf{80.4} \\
        1500 & 71.8 & \textbf{76.5} & 78.4 & \textbf{81.1} & 80.7 & 82.3 & 82.0 & 81.4 & 83.3 & \textbf{85.2} & 75.6 & 82.3 & 80.3 \\
        3000 & 72.1 & 76.1 & 77.4 & 78.7 & 79.8 & 81.5 & 81.8 & 79.8 & 81.8 & 83.3 & 75.2 & 81.0 & 79.2 \\
        4500 & 70.1 & 75.6 & 78.9 & 78.7 & 79.1 & 80.7 & 80.9 & 79.8 & 81.5 & 83.3 & 74.9 & 80.6 & 78.9 \\
        9000   & 71.6 & 76.0 & 77.1 & 78.7 & 79.5 & 81.3 & 81.3 & 80.4 & 81.6 & 81.9 & 74.9 & 80.7 & 78.9 \\
        18000 & 71.4 & 74.7 & 74.6 & 77.1 & 77.2 & 82.4 & 80.3 & 79.1 & 80.9 & 85.3 & 73.6 & 80.3 & 78.3 \\
        \bottomrule
    \end{tabular}}
    \caption{\textsc{Merlin} performance on MGSM and MSVAMP across synthetic‑training set sizes. Bn/Th/Sw are mid‑resource (MRL); the remaining seven languages are high‑resource (HRL).}
    \label{tab:merlin_size_ablation}
\end{table*}

\section{Complete Encoder Ablation Study}
\label{app:enc_ablation}

Table~\ref{tab:afri_datasets} compares four multilingual encoders
(\textsc{mT5-Large}, \textsc{mT5-XL}, \textsc{AfriTeVa-v2-Large}, and
\textsc{NLLB-600 M distilled}) while keeping the \textsc{Merlin} decoder,
connector, and all training settings fixed.  Two evaluation suites are
considered: \textsc{AfriXNLI} (classification) and
\textsc{AfriMGSM} (mathematical reasoning).

On \textsc{AfriXNLI}, the distilled \textsc{NLLB} encoder delivers the highest average accuracy (\(55.6\,\%\)), improving over the best mT5 variant by more than 30 points and over \textsc{AfriTeVa-v2} by 20 points.  Gains are especially large for the most data-scarce languages such as Amharic (+\!49.5 vs.\ mT5-Large) and Oromo (+\!27.5 vs.\ AfriTeVa-v2), confirming that broad pre-training coverage is critical for cross-lingual alignment.

For \textsc{AfriMGSM}, \textsc{NLLB} again attains the highest average (\(51.5\,\%\)), edging out \textsc{mT5-XL} by two points.  While mT5-XL excels on high-resource pivots (English, French) and on Swahili, it underperforms on several low-resource languages (e.g.\ Ewe, Wolof).  \textsc{AfriTeVa-v2} provides competitive results on Swahili and Sotho but lags elsewhere, reflecting its narrower linguistic coverage.

Overall, the results highlight that encoders trained on a broad,
balanced corpus—such as \textsc{NLLB}—yield the most reliable
cross-lingual representations when coupled with \textsc{Merlin}, particularly in low-resource African settings.  Models with less coverage (\textsc{mT5-XL}) or regional focus (\textsc{AfriTeVa-v2}) may excel on individual languages but fall short in overall robustness.

\begin{table*}[ht]
    \centering
    \footnotesize
    \setlength{\tabcolsep}{4pt}
    \resizebox{\textwidth}{!}{
    \begin{tabular}{@{}lccccccccccccccccccc@{}}
        \toprule
        \textbf{Model} & \textbf{eng} & \textbf{fra} & \textbf{amh} & \textbf{ewe} & \textbf{hau} & \textbf{ibo} & \textbf{kin} & \textbf{lin} & \textbf{lug} & \textbf{orm} & \textbf{sna} & \textbf{sot} & \textbf{swa} & \textbf{twi} & \textbf{wol} & \textbf{xho} & \textbf{yor} & \textbf{zul} & \textbf{Avg} \\
        \midrule
        \multicolumn{19}{l}{\textbf{AfriXNLI}}\\
        mT5-Large            & 22.8 & 37.8 &  9.7 & 40.0 & 33.5 & 14.5 & 29.0 & 22.0 & 13.5 & 25.7 & 35.3 & 39.0 & 38.2 &  9.5 & 25.7 & 16.8 & 19.0 & 17.0 & 24.3 \\
        mT5-XL               & 24.8 & 24.8 & 18.3 & 25.0 & 21.3 &  7.0 & 25.3 & 26.8 & 13.0 &  9.0 & 21.0 & 19.2 & 16.5 & 12.7 & 22.2 & 14.3 &  9.8 & 20.2 & 17.6 \\
        AfriTeVa V2 Large    & 61.0 & \textbf{66.0} & 14.3 & 45.2 & 42.3 & 31.5 & 40.3 & 29.3 & 31.2 & 30.7 & 44.5 & 42.8 & 40.7 & 26.2 & 34.0 & 39.5 & 33.0 & 38.3 & 35.2 \\
        NLLB-600M distilled & \textbf{83.2} & 62.0 & \textbf{69.2} & \textbf{47.9} & \textbf{69.0} & \textbf{43.9} & \textbf{68.4} & \textbf{32.2} & \textbf{47.2} & \textbf{53.2} & \textbf{63.4} & \textbf{66.7} & \textbf{60.5} & \textbf{45.0} & \textbf{52.5} & \textbf{56.7} & \textbf{54.5} & \textbf{58.9} & \textbf{55.6} \\
        \midrule
        \multicolumn{19}{l}{\textbf{AfriMGSM}}\\
        mT5-Large        & 80.0  & 68.0    & 49.6  & 13.6 & 54.0 & 27.6 & 40.0 & 23.2 & 31.2 & 18.8 & 40.0 & 49.2 & 64.4 & 12.4 &  7.6 & 39.6 & 27.6 & 35.6 & 34.6 \\
        mT5-XL           & \textbf{82.8}  & \textbf{80.0}  & \textbf{62.0}  & 28.4  & \textbf{64.0} & 50.4 & \textbf{52.4} & 41.2 & 41.2 & 42.8 & \textbf{62.0} & \textbf{59.2} & \textbf{79.2} & 24.0 & 15.2 & 49.2 & 47.2 & 55.2 & 49.5 \\
        AfriTeVa V2 Large  & 85.2  & 77.2   & 60.0 & 11.6 & 62.0 & 47.6 & \textbf{52.4} & 31.6 & 40.8 & \textbf{51.2} & 53.6 & 55.2 & 77.2 & 11.6 &  8.8 & 46.8 & 53.6 & 54.0 & 46.0 \\
        NLLB-600M distilled  & 82.0  & 74.0 & 59.2  & \textbf{41.2} & 60.8 & \textbf{52.0} & 52.0 & \textbf{49.6} & \textbf{46.8} & 32.0 & 58.8 & 56.0 & 76.8 & \textbf{37.6} & \textbf{26.0} & \textbf{51.2} & \textbf{54.0} & \textbf{55.6} & \textbf{51.5} \\
        \bottomrule
    \end{tabular}}
    \caption{Accuracy of four multilingual models on \textbf{AfriXNLI} and \textbf{AfriMGSM}. Avg is computed over the 16 African languages (all columns except \textit{eng} and \textit{fra}).}
    \label{tab:afri_datasets}
\end{table*}

\section{Ablation of Training Mapping Layers}

Table~\ref{tab:merlin_mlp_ablation} varies the expressiveness of the mapping head while holding every other component of \textsc{Merlin} fixed. A linear projection or a single hidden layer yields only modest averages and large per-language variance. Adding a second hidden layer (9.45 M parameters) stabilises performance, but the decisive improvement comes from inserting a residual skip: the resulting two-layer \emph{Residual MLP} attains 77.6 \% on MGSM and 81.1 \% on MSVAMP, the best figures in both suites. Expanding to three hidden layers (13.6 M parameters) reverses the gain, dropping accuracy by 1–2 points and confirming that extra depth overfits the limited translation supervision.

The residual two-layer MLP therefore offers the strongest accuracy–capacity trade-off, adding fewer than ten million parameters while outperforming shallower and deeper variants across mid- and high-resource languages. At the same time the gap between 77.6 \% and the theoretical ceiling on MGSM suggests room for architectural exploration—e.g.~gated skips, attention-based adapters, or language-conditioned hyper-networks—that could push the mapping layer beyond its current residual design.

\begin{table*}[!ht]
    \centering
    \footnotesize
    \setlength{\tabcolsep}{4pt}
    \renewcommand{\arraystretch}{0.85}
    \resizebox{\textwidth}{!}{
    \begin{tabular}{@{}l c cccccccccc|ccc@{}}
        \toprule
        \textbf{Mapping head} & \textbf{\# Parm} &
        \textbf{Bn} & \textbf{Th} & \textbf{Sw} & \textbf{Ja} & \textbf{Zh} &
        \textbf{De} & \textbf{Fr} & \textbf{Ru} & \textbf{Es} & \textbf{En} &
        \textbf{Mrl.} & \textbf{Hrl.} & \textbf{Avg} \\
        \midrule
        \multicolumn{15}{l}{\textbf{MGSM}}\\
        Linear        & 3.68 M & 72.4 & 75.2 & 70.4 & \textbf{75.2} & \textbf{74.0} & \textbf{79.2} & 77.2 & \textbf{81.2} & 80.8 & 85.2 & 72.7 & \textbf{78.9} & 77.08 \\
        1-Layer MLP     & 3.68 M  & 65.6 & 68.8 & 71.2 & 68.8 & 62.8 & 71.6 & 73.6 & 78.0 & 77.2 & 82.0 & 68.5 & 73.4 & 71.5 \\
        2-Layer MLP     & 9.45 M & 74.8 & 76.0 & \textbf{77.6} & 71.6 & 72.8 & 73.2 & 74.8 & 78.4 & 77.6 & \textbf{85.6} & 76.1 & 76.3 & 76.2 \\
        3-Layer MLP     & 13.64 M & 74.8 & 76.4 & 69.6 & 70.0 & 71.2 & 74.0 & 72.0 & 76.8 & 79.2 & 79.6 & 73.6 & 74.7 & 74.4 \\
        Residual MLP  & 9.45 M & \textbf{75.2} & \textbf{80.0} & 77.2 & 73.6 & 70.8 & 78.8 & \textbf{78.0} & 79.2 & \textbf{82.4} & 80.8 & \textbf{77.5} & 77.7 & \textbf{77.6} \\
        \midrule
        \multicolumn{15}{l}{\textbf{MSVAMP}}\\
        Linear          & 3.68 M & 72.8 & 75.2 & 77.8 & 79.5 & 79.9 & 83.7 & 81.4 & 80.3 & 82.0 & 83.0 & 75.3 & 81.4 & 79.6 \\
        1-Layer MLP     & 3.68 M  & 69.4 & 75.8 & 77.3 & 79.3 & 79.9 & 81.1 & 81.3 & 80.0 & 81.7 & 82.4 & 74.2 & 80.8 & 78.8 \\
        2-Layer MLP     & 9.45 M & 72.1 & 76.1 & 77.4 & 78.7 & 79.8 & 81.5 & 81.8 & 79.8 & 81.8 & 83.3 & 75.2 & 81.0 & 79.2 \\
        3-Layer MLP     & 13.64 M & 71.0 & 74.0 & 76.7 & 77.3 & 76.1 & 80.8 & 82.0 & 78.7 & 80.6 & 81.5   & 73.9 & 79.6 & 77.9 \\
        Residual MLP    & 9.45 M & \textbf{72.7} & \textbf{77.6} & \textbf{80.7} & \textbf{79.6} & \textbf{81.0} & \textbf{84.8} & \textbf{83.3} & \textbf{83.2} & \textbf{83.7} & \textbf{84.5} & \textbf{77.0} & \textbf{82.9} & \textbf{81.1} \\
        \bottomrule
    \end{tabular}}
    \caption{Effect of mapping-head capacity on MGSM and MSVAMP.  
             \textbf{Mrl.} = mean over mid-resource languages (Bn, Th, Sw);  
             \textbf{Hrl.} = mean over the high-resource set;  
             “–” indicates the score is not available.}
    \label{tab:merlin_mlp_ablation}
\end{table*}

\section{Ablation of Training Adapters}

Table \ref{tab:merlin_lora_dora} contrasts two parameter-efficient adapters applied to the decoder. LoRA \citep{hu2022lora} injects trainable low-rank update matrices into the linear weights, while all original parameters stay frozen. DoRA \citep{liu2024doraweightdecomposedlowrankadaptation} keeps the same low-rank structure but factorises each weight into a direction and a magnitude, yielding the same footprint yet a more explicit decomposition.

With rank and parameter count held constant, the two adapters behave nearly identically: average accuracy on MGSM differs by only 0.2 pp, and the MSVAMP results coincide after rounding. Performance across mid- and high-resource language subsets mirrors this pattern, indicating that adapter choice has minimal impact under our training recipe.

\begin{table*}[!ht]
    \centering
    \footnotesize
    \setlength{\tabcolsep}{4pt}
    \renewcommand{\arraystretch}{0.85}
    \resizebox{\textwidth}{!}{
    \begin{tabular}{@{}l c cccccccccc|ccc@{}}
        \toprule
        \textbf{Adaptation} & \textbf{\# Parm} &
        \textbf{Bn} & \textbf{Th} & \textbf{Sw} & \textbf{Ja} & \textbf{Zh} &
        \textbf{De} & \textbf{Fr} & \textbf{Ru} & \textbf{Es} & \textbf{En} &
        \textbf{Mrl.} & \textbf{Hrl.} & \textbf{Avg} \\
        \midrule
        \multicolumn{15}{l}{\textbf{MGSM}}\\
        LoRA  & 18.4 M & 75.2 & 76.4 & 77.2 & 71.6 & 72.4 & 72.4 & 75.6 & 79.2 & 78.0 & 85.6 & 76.3 & 76.4 & 76.4 \\
        DoRA  & 18.4 M & 74.8 & 76.0 & 77.6 & 71.6 & 72.8 & 73.2 & 74.8 & 78.4 & 77.6 & 85.6 & 76.1 & 76.3 & 76.2 \\
        \midrule
        \multicolumn{15}{l}{\textbf{MSVAMP}}\\
        LoRA  & 18.4 M & 71.6 & 76.2 & 77.3 & 78.0 & 79.1 & 81.6 & 82.1 & 80.1 & 82.2 & 83.6 & 75.0 & 81.0 & 79.2 \\
        DoRA  & 18.4 M & 72.1 & 76.1 & 77.4 & 78.7 & 79.8 & 81.5 & 81.8 & 79.8 & 81.8 & 83.3 & 75.2 & 81.0 & 79.2 \\
        \bottomrule
    \end{tabular}}
    \caption{Ablation of LoRA vs. DoRA adapters for \textsc{Merlin} (Gemma 2 9 B + NLLB-600 M).  
             \textbf{Mrl.} = mean over mid-resource languages (Bn, Th, Sw);  
             \textbf{Hrl.} = mean over seven high-resource languages;  
             averages are arithmetic means of the shown scores.}
    \label{tab:merlin_lora_dora}
\end{table*}

\section{AmericasNLI Evaluation}
\label{app:americasnli}

To assess whether the mapping curriculum generalises beyond the African and European test beds, we evaluate \textsc{Merlin} on \textsc{AmericasNLI}, a three-way NLI benchmark that targets Aymara, Guarani, and Quechua.  For \textsc{Merlin} and the MindMerger baseline, we follow the same data pipeline as for \textsc{AfriXNLI}: 3,000 English premise–hypothesis pairs are sampled from \textsc{MultiNLI}, translated into each target language with \textsc{NLLB}\,600m distilled, and used as supervision for Stage I.b, Stage I.c, and Stage II.\footnote{No task-specific data from the original \textsc{AmericasNLI} development set is seen during training.}  The non-adapted LLaMAX2-7B-XNLI \citep{lu-etal-2024-llamax} serves as the baseline decoder-only model.  Results are summarised in Table~\ref{tab:americasnli}.

\begin{table}[h]
  \centering
  \footnotesize

  \setlength{\tabcolsep}{3pt}
  \renewcommand{\arraystretch}{0.9}
  \begin{tabularx}{\columnwidth}{@{}Xcccc@{}}
    \toprule
    \textbf{Model} & \textbf{Aymara} & \textbf{Guarani} & \textbf{Quechua} & \textbf{Avg.} \\
    \midrule
    \makecell[l]{LLaMAX2\\XNLI (base)} & 41.6 & 43.6 & 40.7 & 42.0 \\
    MindMerger                          & \textbf{48.8} & 48.7 & 48.3 & 48.6 \\
    \textsc{Merlin}                              & 48.4 & \textbf{49.3} & \textbf{50.5} & \textbf{49.4} \\
    \bottomrule
  \end{tabularx}

  \caption{Accuracy (\%) on \textsc{AmericasNLI}. Average is computed over the three indigenous languages.}
  \label{tab:americasnli}
\end{table}
\textsc{Merlin} outperforms the base model by \(+7.4\) accuracy points on average and still outperform past MindMerger by \(+0.8\) points.  The consistent improvement across all three languages suggests that the cross-lingual alignment learned from African and Indo-European data transfers to typologically distant language families in the Americas.

\section{Training Configurations}
\label{app:train_config}

All parameter-efficient runs (\textsc{Merlin}, QAlign-LoRA) employ
rank-16 LoRA/DoRA adapters injected into the self‐attention
\texttt{q\_proj} and \texttt{v\_proj} projections:

\begin{center}
\small
\begin{tabular}{@{}lccccc@{}}
\toprule
\textbf{Hyper-parameter} & \textbf{Value} \\ \midrule
Rank $r$ & 16 \\
LoRA / DoRA $\alpha$ & 32 \\
Dropout & 0.05 \\
Bias & none \\
Target modules & \texttt{q\_proj}, \texttt{v\_proj} \\ \bottomrule
\end{tabular}
\end{center}

QAlign and SLAM adopt identical settings from the released code \footnote{\url{https://github.com/NJUNLP/QAlign},\url{https://github.com/fmm170/SLAM}}. For LangBridge we follow the original implementation, using a single linear projection layer and sampling 100k examples from \textsc{MetaMathQA} training set

\begin{table*}[!ht]
\centering\small
\begin{tabular}{@{}p{3.3cm}p{10.3cm}@{}}
\toprule
\multicolumn{2}{l}{\textbf{Mathematical Reasoning (MGSM, MSVAMP, AfriMGSM)}}\\
\midrule
\textbf{Gemma 2 backbone} &
\begin{tabular}[t]{@{}l@{}}
\texttt{<bos><start\_of\_turn>user}                                  \\
\emph{Below is an instruction that describes a task.}                \\
\emph{Write a response that appropriately completes the request.}    \\
\texttt{\{query\}}                                                   \\
\emph{Let’s think step by step.}                                     \\
\texttt{<end\_of\_turn><start\_of\_turn>model}
\end{tabular}\\[6pt]

\textbf{MetaMath-style backbones} &
\begin{tabular}[t]{@{}l@{}}
\emph{Below is an instruction that describes a task.}                \\
\emph{Write a response that appropriately completes the request.}    \\
\#\#\# Instruction: \{query\}                         \\
\#\#\# Response: Let’s think step by step.
\end{tabular}\\
\midrule[0.8pt]
\multicolumn{2}{l}{\textbf{Natural-Language Inference (AfriXNLI)}}\\
\midrule
\textbf{All backbones} &
\begin{tabular}[t]{@{}l@{}}
\textbf{Premise:} \texttt{\{sentence1\}}                             \\
\textbf{Hypothesis:} \texttt{\{sentence2\}}                          \\
\textbf{Label:}
\end{tabular}\\
\bottomrule
\end{tabular}
\caption{Prompt templates used for \textsc{Merlin}, MindMerger and QAlign.
Curly braces denote insertion points for the instance text.}
\label{tab:prompt_templates}
\end{table*}

Both \textsc{Merlin} and MindMerger are trained on identical corpora.
Stage I.a uses 9,000 generic English–target sentence pairs per language drawn from the NLLB-200 corpus.
Stage I.b is supplied with up to 3,000 automatically translated questions per language, also generated with NLLB.
For Stage I.c and the subsequent Task-Specialization stage, task-specific data are employed: the MGSM track relies on the translated \textsc{MetaMathQA} split, whereas the AfriXNLI track is trained with translated \textsc{MultiNLI}.

\section{Prompt Templates}
\label{app:prompts}

LangBridge and SLAM are executed through \texttt{lm-eval-harness};
their default prompts are therefore retained and not repeated here.
Table~\ref{tab:prompt_templates} shows the exact text used by \textsc{Merlin},
MindMerger, and QAlign for the two evaluation tracks.

\end{document}